\documentclass[11pt]{article}
\PassOptionsToPackage{square,numbers}{natbib}
\usepackage{emnlp2016}
\emnlpfinalcopy
\usepackage{url}
\usepackage{latexsym}
\usepackage{amsmath}
\usepackage{amssymb}
\usepackage{color,soul}
\usepackage{booktabs}
\usepackage{graphicx}
\usepackage{lipsum}
\usepackage{nicefrac}       
\usepackage{microtype}      
\usepackage{natbib}
\usepackage{subfig}

\usepackage{tikz}
\usetikzlibrary{calc,trees,positioning,arrows,chains,shapes.geometric,%
  decorations.pathreplacing,decorations.pathmorphing,shapes,%
  matrix,shapes.symbols,fit,decorations}
\usepackage{wasysym}
\usepackage{xcolor}
\definecolor{nice-red}{HTML}{E41A1C}
\definecolor{nice-orange}{HTML}{FF7F00}
\definecolor{nice-yellow}{HTML}{FFC020}
\definecolor{nice-green}{HTML}{4DAF4A}
\definecolor{nice-blue}{HTML}{377EB8}
\definecolor{nice-purple}{HTML}{984EA3}

\newcommand{\eg}{{\emph{e.g.}}}

\title{MuFuRU: The Multi-Function Recurrent Unit}

\author{Dirk Weissenborn\\
	    Language Technology Lab, DFKI\\
	    Alt-Moabit 91c\\
	    Berlin, Germany\\
        {\tt dirk.weissenborn@dfki.de} \\ 
        \And
        Tim Rockt\"aschel\\
	    University College London\\
	    Gower Street\\
	    London, UK\\
        {\tt t.rocktaschel@cs.ucl.ac.uk} \\}

\date{}

\begin{document}
\maketitle
\begin{abstract}
Recurrent neural networks such as the GRU and LSTM found wide adoption in natural language processing and achieve state-of-the-art results for many tasks. 
These models are characterized by a memory state that can be written to and read from by applying gated composition operations to the current input and the previous state. 
However, they only cover a small subset of potentially useful compositions. 
We propose Multi-Function Recurrent Units (MuFuRUs) that allow for arbitrary differentiable functions as composition operations. 
Furthermore, MuFuRUs allow for an input- and state-dependent choice of these composition operations that is learned. Our experiments demonstrate that the additional functionality helps in different sequence modeling tasks, including the evaluation of propositional logic formulae, language modeling and sentiment analysis.
\end{abstract}

\section{Introduction}

Recurrent neural networks (RNNs) have been applied successfully to a great variety of sequence modeling tasks. Impressive results were achieved for tasks that involve text, such as language modeling \cite{mikolov2010recurrent}, machine translation \cite{sutskever2014sequence}, sentiment analysis \cite{tai2015improved}, document-level question answering \cite{nips15_hermann} or recognizing textual entailment \cite{rocktaschel2015reasoning}, to name just a few. Modern architectures extend RNNs with additional functionality like attention \cite{bahdanau2014neural} or external memory \cite{graves2014neural,sukhbaatar2015end}.

At the core of every RNN is a recurrent \textit{cell}-function that specifies how the current input and the previous state should be combined to form a new state. Different cell-functions have been proposed in the past, including the traditional $\tanh$-cell (Vanilla), the Long-Short-Term-Memory (LSTM \cite{hochreiter1997long}) or more recently the Gated Recurrent Unit (GRU \cite{chung2014empirical}). They allow for either replacing (Vanilla, GRU, LSTM), keeping (GRU, LSTM) or additively aggregating (LSTM) features in every hidden dimension. The decision is realized via soft gating mechanisms.

Though different extensions and variations to GRUs and LSTMs have been investigated recently \cite{greff2015lstm,jozefowicz2015empirical}, none of them outperform standard GRUs or LSTMs significantly on a range of different tasks. 
We believe a promising direction towards a more task-adaptive RNN architecture is to (i) allow for other differentiable composition operations and (ii) learn input-dependent selection of these operations end-to-end from task data.

Therefore, we propose a novel \textit{cell}-function, called \textit{Multi-Function Recurrent Unit} (MuFuRU). It is a generalization of frequently-used recurrent architectures and allows for the introduction of arbitrary differentiable composition operations of the previous RNN state and newly constructed feature vector. Crucially, the selection of the composition operation itself is learned and dependent on the previous state and current input. Thus the function that is applied to obtain the next state can change while processing inputs.

Our contributions are threefold: (i) we discuss existing commonly-used RNN architectures and introduce the Multiple-Function Recurrent Unit (Sections \ref{sec:rnn} and \ref{sec:mufuru}), (ii) we demonstrate that MuFuRUs can learn to evaluate simple logical expressions with comparably small memory size and without over-fitting when the memory size is increased (Section \ref{sec:logic}), and 
(iii) we show that MuFuRU outperforms a standard GRU baseline on language modeling and sentiment analysis (Sections \ref{sec:lm} and \ref{sec:sentiment}) where it approaches state-of-the-art results without hyper-parameter tuning.

\section{Recurrent Neural Networks}
\label{sec:rnn}

A recurrent neural network is fully specified via a recurrent function $f_{\boldsymbol{\theta}}: \mathbb{R}^N \times \mathbb{R}^M \to \mathbb{R}^M \times \mathbb{R}^H$ parameterized by $\boldsymbol{\theta}$, where $M$ is the size of the state vector, $H$ the size of the output vector and $N$ the size of the input vector.
Given an input sequence $\boldsymbol{X}=(\boldsymbol{x}_1, ..., \boldsymbol{x}_T)$ and a start state $\boldsymbol{s}_0$, the state and output at time step $t \in [1,T]$ is computed as  $(\boldsymbol{s}_t,\boldsymbol{h}_t)=f(\boldsymbol{x}_t, \boldsymbol{s}_{t-1})$ . Finally, an RNN is defined as the recurrent application of the cell-function $f$ to inputs and previous states from time step $1$ to $T$.
\paragraph{Vanilla RNN} The simplest cell-function is the $\tanh$-cell, where the new state at each time step is computed by a non-linear projection of the current input $\boldsymbol{x}_t$ and the previous state. Note, that the output of the $\tanh$-cell is its state.

\paragraph{Gated Recurrent Unit} The GRU updates its state by the element-wise, weighted sum of a newly constructed feature vector $\boldsymbol{v}_t$ and the previous state $\boldsymbol{s}_{t-1}$ via the update gate $\boldsymbol{u}_t$. The reset gate $\boldsymbol{r}_t$ selects which features of the previous hidden state will be used to create the new features. This is useful in situations where the previous state should be forgotten in favor of the creation of new features. Eq.~\ref{eq:gru} shows how $\boldsymbol{h}_t$ and $\boldsymbol{s}_t$ are computed at each time step.

\begin{align}
	\begin{bmatrix}
	\boldsymbol{r}_t \\
	\boldsymbol{u}_t
	\end{bmatrix} &= 
	\sigma\left( W_u \begin{bmatrix} \boldsymbol{x}_t \\ \boldsymbol{s}_{t-1} \end{bmatrix} + \boldsymbol{b}_u \right)  \nonumber\\ 
	\boldsymbol{v}_t &= \tanh \left( W_v \begin{bmatrix} \boldsymbol{x}_t \\ \boldsymbol{r}_t \odot \boldsymbol{s}_{t-1} \end{bmatrix} + \boldsymbol{b}_v \right) \nonumber \\
	\boldsymbol{s}_t &= \boldsymbol{h}_t = \boldsymbol{u}_t \odot \boldsymbol{s}_{t-1}+ (1-\boldsymbol{u}_t) \odot \boldsymbol{v}_t \nonumber \label{eq:gru} \\
\end{align} 

\section{Multi-Function Recurrent Unit}
\label{sec:mufuru}
The Multi-Function Recurrent Unit (MuFuRU) applies a predefined set of composition operations (such as element-wise max, min, absolute difference etc.) to a new feature vector and the previous state, and decides which composition should be used for every feature dimension individually.

\subsection{Architecture}

At every time step $t$ the MuFuRU calculates normalized weights $\boldsymbol{p}_t^{j}$ for all composition operations $\operatorname{op}^j: \mathbb{R}^M \times \mathbb{R}^M \to \mathbb{R}^M$,~$j \in [1,l]$, via an operation controller $\boldsymbol{k}_t$ (Equation \ref{eq:weights}).
\begin{align}
	\boldsymbol{\hat{p}}_t^j &= W_{p}^j \boldsymbol{k}_t + \boldsymbol{b}_{p}^j  \nonumber\\ 
	\left[\boldsymbol{p}_t^1,... , \boldsymbol{p}_t^{l} \right] &= \operatorname{softmax} \left( \left[\boldsymbol{\hat{p}}_t^1,... , \boldsymbol{\hat{p}}_t^{l} \right] \right) \label{eq:weights}
\end{align}
In this work, the operation controller is the concatenation of $\boldsymbol{s}_{t-1}$ and the current input $\boldsymbol{x}_t$ (Equation~\ref{eq:controller}).\footnote{It is also possible to compute a lower dimensional, recurrent controller at every time step to save parameters.}

\begin{align}
\boldsymbol{k}_t&=\begin{bmatrix} \boldsymbol{x}_t \\ \boldsymbol{s}_{t-1} \end{bmatrix} \label{eq:controller} 
\end{align}

The MuFuRU (like the GRU) utilizes a reset-gate for the computation of the new feature vector $\boldsymbol{v}_t$ and combines it with the previous hidden state by a convex combination of the $l$ different composition operations $\operatorname{op}^j$ (Equation~\ref{eq:MuFuRU}).

\begin{align}
	\boldsymbol{r}_t  &= \sigma\left( W_r \begin{bmatrix} \boldsymbol{x}_t \\ \boldsymbol{s}_{t-1} \end{bmatrix} + \boldsymbol{b}_r \right)  \nonumber \\ 
	\boldsymbol{v}_t &= \tanh \left( W_v \begin{bmatrix} \boldsymbol{x}_t \\ \boldsymbol{r}_t \odot \boldsymbol{s}_{t-1} \end{bmatrix} + \boldsymbol{b}_v \right) \nonumber \\
	\boldsymbol{s}_t &= \boldsymbol{h}_t = \sum_{j=1}^{l} \boldsymbol{p}_t^j \cdot \operatorname{op}^j(\boldsymbol{s}_{t-1}, \boldsymbol{v}_{t} )  \label{eq:MuFuRU}
\end{align}
In summary, a MuFuRU learns to select the composition function from a predefined set of operations based on the current input and previous state.

\subsection{Composition Functions}
In Table \ref{tab:ops} we list the composition functions that were used in this work. Note that this list can easily be extended by other differentiable functions suitable for a given task. The input to each operation is the previous state $\boldsymbol{s}$ and a feature vector $\boldsymbol{v}$.
\begin{table}[t!]
\centering
\resizebox{0.5\textwidth}{!}{
    \begin{tabular}{ll}
    \toprule
    Operator & Description\\
    \midrule
    $\operatorname{keep}(\boldsymbol{s}, \boldsymbol{v}) = \boldsymbol{s}$ & Keep previous hidden state\\
    $\operatorname{replace}(\boldsymbol{s}, \boldsymbol{v}) = \boldsymbol{v}$ & Replace state with new features\\
    $\operatorname{max}(\boldsymbol{s}, \boldsymbol{v})$ & Maximum in every dimension\\
    $\operatorname{min}(\boldsymbol{s}, \boldsymbol{v})$ & Minimum in every dimension\\
    $\operatorname{mul}(\boldsymbol{s}, \boldsymbol{v}) = \boldsymbol{s} \odot \boldsymbol{v}$ & Element-wise multiplication\\
    $\operatorname{diff}(\boldsymbol{s}, \boldsymbol{v}) = 0.5 \cdot |\boldsymbol{s} - \boldsymbol{v}|$ & Element-wise absolute difference\\
    $\operatorname{forget}(\boldsymbol{s}, \boldsymbol{v}) = \boldsymbol{0}$ & Reset current state to zero\\
    \bottomrule
    \end{tabular}
}    
    \caption{MuFuRU operations used in this work.}
    \label{tab:ops}
\end{table}
   
\subsection{Relations to existing Architectures}

The MuFuRU is a generalization of existing architectures. Therefore, with slight adaptations and a correct choice of composition functions, the MuFuRU becomes the same or similar to existing architectures.

\paragraph{Vanilla RNN} The MuFuRU becomes a standard $\tanh$-cell when only the $\operatorname{replace}$-operation is allowed and the reset-gate is set to $\boldsymbol{1}$.

\paragraph{GRU} A GRU is a MuFuRU where only the $\operatorname{keep}$- and $\operatorname{replace}$-operation are allowed.

It follows that the MuFuRU should in principle be able to perform at least as well on sequence modeling tasks as those architectures, since it can learn to operate like these.

\section{Experiments}

We evaluate the MuFuRU with the composition operations shown in Table \ref{tab:ops} on different tasks that involve modeling of sequences. 
In the following experiments we always compare performance to the GRU as baseline. 
Every model is trained with the same task-specific hyper-parameters to ensure comparability. Although the MuFuRU via the operation controller contains a larger parameter set than a GRU, we believe that this does not affect comparability since the additional parameters are only concerned with selecting a single operation at each time step and our goal is to investigate whether the introduction of new operations is beneficial or not. Besides, all models easily overfit in most of our experiments giving an advantage to models with better generalization ability.

We perform mini-batch stochastic gradient descent using ADAM \cite{kingma2014adam} with $\beta_1=0.0$ (no momentum) and $\beta_2=0.999$ for optimization in all experiments. 

\subsection{Propositional Logic}
\label{sec:logic}

\begin{figure*}[ht!]
     \centering
     \subfloat[b][Accuracy on the propositional logic test set for GRU and MuFuRU with different hidden dimensions.]{\includegraphics[width=0.37\textwidth]{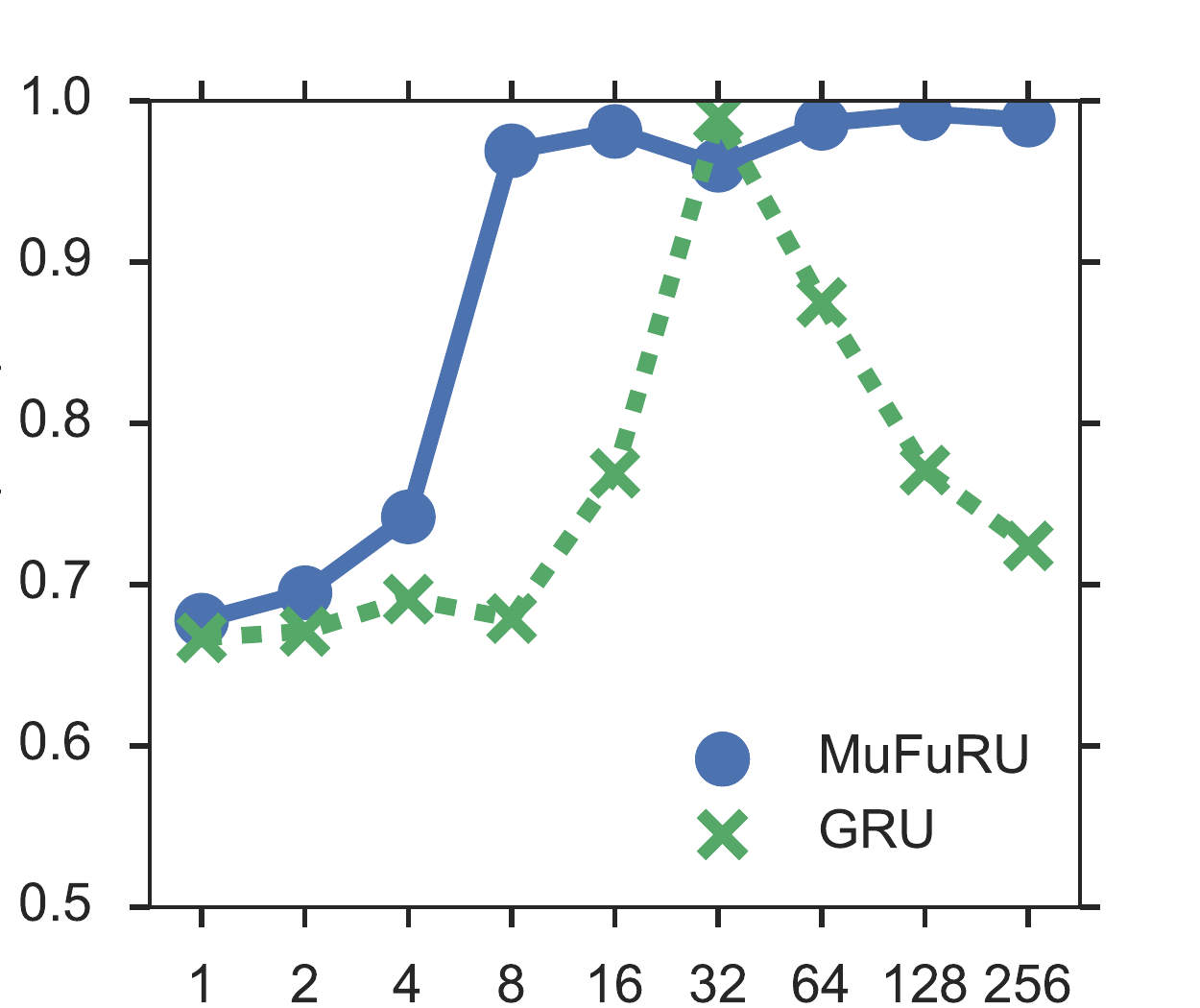}\label{fig:logic_results}}
     \hfill
     \subfloat[b][Average weight of operations in the propositional logic experiment.]{\includegraphics[width=0.34\textwidth]{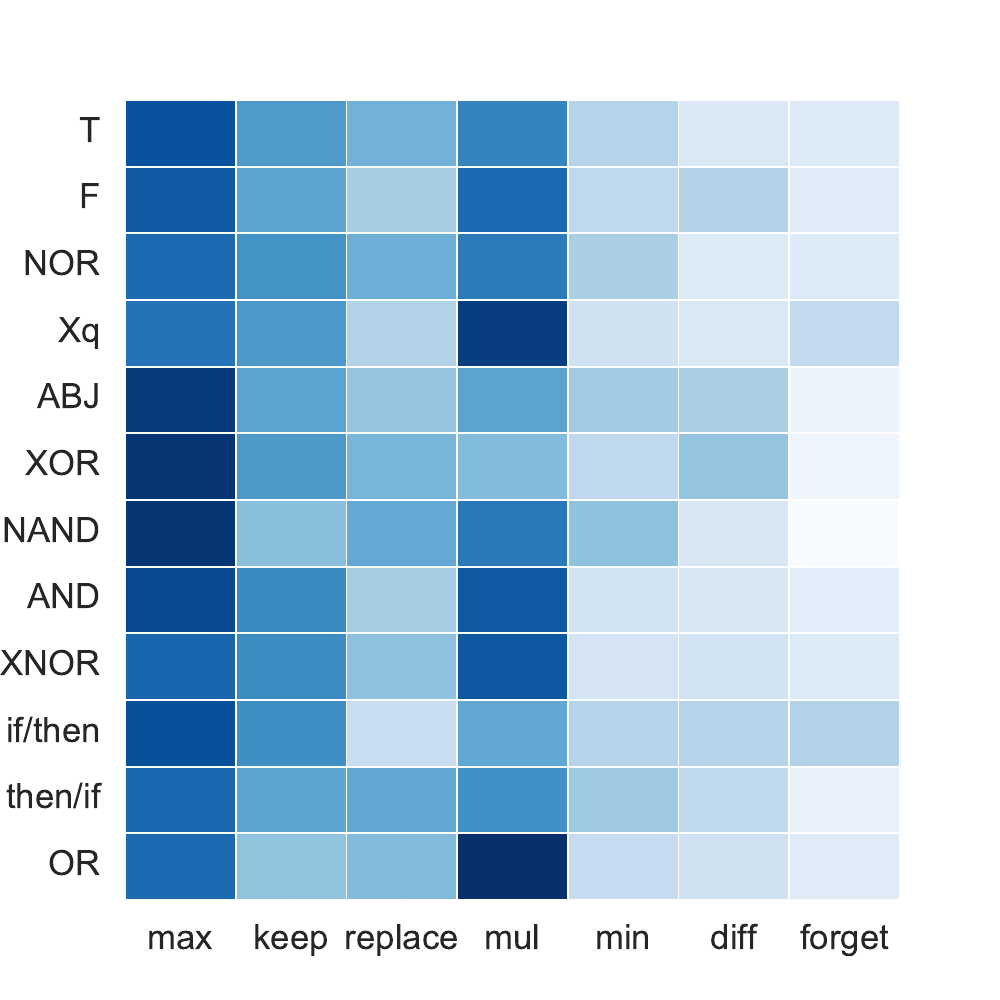}\label{fig:op_weights}}
     \hfill
     \subfloat[b][Results for binary subtask of the Sentiment Treebank.]{%
        \resizebox{0.25\textwidth}{!}{%
            \begin{tabular}[b]{l c}
                \toprule
                \textbf{Model} & \textbf{Accuracy} \\ 
                \midrule
                RNTN \cite{socher2013recursive} & $85.4$ \\
                DCNN \cite{blunsom2014convolutional} & $86.8$ \\
                CNN-MC \cite{kim2014convolutional} & $\mathbf{88.1}$ \\
                CT-LSTM \cite{tai2015improved} & $88.0$ \\
                LSTM \cite{tai2015improved} & $84.9$ \\
                \midrule
                GRU & $87.3$ \\
                MuFuRU & $\underline{87.6}$ \\
                \bottomrule\\
            \end{tabular}
        }
        \label{tab:sst_results}
     }
     \caption{Results for propositional logic (a,b) and sentiment analysis (c) tasks.}
     \label{results}
\end{figure*}
As a unit test we evaluate the MuFuRU's ability to learn to evaluate simple propositional logic formulae. We sample sequences of Boolean binary gates and input truth values. For instance $[1, 0, \wedge, 0, \vee, 1, \Rightarrow]$ represents the Boolean expression $((1 \wedge 0) \vee 0) \Rightarrow 1$. We train the GRU and MuFuRU on 1000 Boolean formulae with 5-10 gates and test on 1000 unseen longer formulae with 11-20 gates.

Figure \ref{fig:logic_results} shows the test accuracy of a GRU and MuFuRU with different hidden dimensions trained for 100 epochs. While the GRU struggles to generalize to longer sequences, the MuFuRU learns to evaluate Boolean formulae with a memory size of only 8. The GRU can emulate the operations needed to evaluate Boolean gates when provided with a much larger hidden dimension, but it quickly starts to overfit with larger hidden dimensions. In contrast, the MuFuRU does not overfit even for large hidden dimensions, which indicates that it learns to apply the correct arithmetic counterpart for every Boolean operation.

We plot the average weight of operations used in the MuFuRU for input truth values and Boolean gates in Figure \ref{fig:op_weights}\footnote{See \cite{graves2016adaptive} for a description of the gates.}. Some weights seem counter-intuitive (\eg{}, multiplication has the highest weight for modeling OR). However, note that the choice of operation is dependent not only on the current input but also the previous state (see Eq. \ref{eq:MuFuRU}). We find that the MuFuRU learns a specific behavior tailored to different inputs and that it makes sparse use of the provided operations.

\subsection{Language Modeling}
\label{sec:lm}

Language Modeling is an important task for all applications involving language generation. It requires a system to predict the next word conditioned on the previous text at every time step. For this experiment we use the PTB dataset \cite{mikolov2010recurrent} with a limited vocabulary of 10k words. We trained single-layer models with $200$ hidden units. The GRU with a testset perplexity of $123.0$ was outperformed by the MuFuRU with a perplexity of $\mathbf{119.7}$. 
This result substantiates our claim that the MuFuRU should be at least as good at modeling sequences as the GRU.

\subsection{Sentiment Analysis}
\label{sec:sentiment}
Sentiment classification requires a system to recognize the polarity of a text. 
In this experiment, we train our models on the Sentiment Treebank \cite{socher2013recursive}, which contains 215,154 annotated phrases collected from 11,855 sentences. 

We trained the models with $100$ hidden units using mini-batches of 25 phrases each and feed the final output vector at the end of each phrase as input to a logistic regression classifier. Word embeddings are tuned and initialized with \verb|Glove| \cite{pennington2014glove} or sampled uniformly between -0.05 and 0.05 for unknown words. We chose the model with the best accuracy on the development set in each run which was evaluated every 200 mini-batches.

The results of our experiments along with current state-of-the-art results are presented in Figure~\ref{tab:sst_results}. Both, the GRU and MuFuRU, achieve high accuracy. The MuFuRU performs better than the GRU, which indicates that the introduction of new operations helps in this task. The MuFuRU even approaches state-of-the-art results without the need for complex structural biases. However, using more complex structures for RNNs is orthogonal to this work, as the differentiable operations of the MuFuRU can be integrated into other architectures.

\section{Conclusion}

We presented the \textit{Multi-Function Recurrent Unit} (MuFuRU), a new recurrent neural network architecture that learns to select composition functions for the combination of computed features with an existing state at every time step. It thereby generalizes beyond existing models that are limited to a very small set of such compositions. We demonstrate its theoretical advantages on a toy task that evaluates simple propositional formulae and provide empirical evidence on a language modeling task that additional compositional functionality is useful. Since MuFuRUs are in principle able to learn the same or similar behaviour as GRUs or LSTMs, they can be used in place of these \textit{cell}-functions in RNNs. 
Furthermore, other task-specific differentiable composition functions can easily be integrated.

\section*{Acknowledgments}
This research was partially supported by Microsoft Research through its PhD Scholarship Programme and by the German Federal Ministry of Education and 
Research (BMBF) through the projects ALL SIDES (01IW14002), BBDC (01IS14013E), and Software Campus (01IS12050, sub-project GeNIE).

\bibliography{mufuru}
\bibliographystyle{abbrvnat}

\appendix

\end{document}